\documentclass[conference,compsoc]{IEEEtran}
\ifCLASSOPTIONcompsoc
  \usepackage[nocompress]{cite}
\else
  \usepackage{cite}
\fi
\usepackage{amsmath}
\usepackage{bm}
\usepackage{cuted}
\usepackage{graphicx}
\ifCLASSINFOpdf
\else
\fi
\begin{document}
%
\title{Leveraging Reinforcement Learning for evaluating Robustness of KNN Search Algorithms}

\author{
\IEEEauthorblockN{Pramod Vadiraja}
\IEEEauthorblockA{TU Kaiserslautern\\
67663 Kaiserslautern\\
Email: p\_vadiraja18@cs.uni-kl.de}

\and

\IEEEauthorblockN{Christoph Peter Balada}
\IEEEauthorblockA{German Research Center for Artificial Intelligence\\
67663 Kaiserslautern\\
Email: Christoph\_Peter.Balada@dfki.de}

}


%


\maketitle

\begin{abstract}
The problem of finding K-nearest neighbors in the given dataset for a given query point has been worked upon since several years. In very high dimensional spaces the K-nearest neighbor search (KNNS) suffers in terms of complexity in computation of high dimensional distances. With the issue of curse of dimensionality, it gets quite tedious to reliably bank on the results of variety approximate nearest neighbor search approaches. In this paper, we survey some novel K-Nearest Neighbor Search approaches that tackles the problem of Search from the perspectives of computations, accuracy of approximated results and leveraging parallelism to speed-up computations. We attempt to derive a relationship between the true positive and false points for a given KNNS approach. Finally, in order to evaluate the robustness of a KNNS approach against adversarial points, we propose a generic Reinforcement Learning based framework for the same.
\end{abstract}


%
\IEEEpeerreviewmaketitle

\section{Introduction}
The K-Nearest Neighbor Search (KNNS) problem is the task of finding neighboring points in the given data set that is closest to the specified query point. Specifically it deals with finding K-nearest points to the query point. The results obtained from this problem can in-turn be used for a classification or a regression problem. In case of classification the query point is assigned a label that is in majority among its K-nearest points. In case of regression, the output with respect to the query point could be an average value of the K-nearest points or a more complicated function of the nearest points. More generally the points belong to a metric space and the dissimilarity between the points is expressed as a distance metric. The problem of KNNS, despite its simplicity has been a go-to technique for classification and regression problems in areas like finance, agriculture \cite{predicting_events}, data transformation processes for businesses, in manufacturing to perform deeper analysis on machines where there is a requirement of exact readings of the machine parameters \cite{businesses}. One of the application from Facebook AI Research (FAIR) \cite{khandelwal2019generalization} include interpolating the state-of-the-art pre-trained language model with the KNNS to predict the next nearest word where they realised KNNS might be effective for language modeling for long-tail patterns.  
Various solutions to the KNNS problem have been proposed. One of the categories include exact algorithms to search for exact neighbors without compromising on the accuracy of the neighbors. The correctness in such approaches is very high. Linear search \cite{KNN_ref}, space partitioning approaches such as BallTree \cite{BallTree}, KD-Tree \cite{KDTree}, Rtree\cite{RTree} are some examples that fall under this category. However, exact approaches are computationally expensive for large number of items where the algorithm needs to perform comparisons with all the points in the data space for every query point. The other suite of algorithms fall under the approximate approaches where the algorithms focus on the fetching the approximate K neighbors. The notion of approximation comes into the picture when the algorithms enforce some amount of relaxation in the distance of the query point to its nearest points. Locality Sensitive Hashing \cite{LSH}, FAANG \cite{FAANG}. EFANNA \cite{EFANNA} are some examples. The efficiency of these approaches is measured in terms of query execution times, distance computation times, the precision and recall of the neighbors of the query point etc.\\
Along with other algorithms, KNNS as well suffers from an issue called the Curse of Dimensionality \cite{CurseOfDim}, that occurs when organizing and analyzing data points in very high dimensional spaces causing various unexpected phenomena. In very high dimensional spaces, the volume of the space increases exponentially such that the available data becomes sparse. Hence, it becomes tedious to group, organize and categorise points in such spaces. The points in high dimensional spaces might get dispersed from each other or the points might end up roughly at the same distance from each other. Such a condition needs to be dealt with in order to avoid erroneous results. Also, it is common to use the Euclidean distance to measure distances between points in very high dimensional spaces. However, there has been experiments \cite{aggarwal2001surprising} that suggests certain key loop-holes in using metrics such as the Euclidean norm in very high dimensional spaces. They suggest that the ratio of the distances of the nearest and farthest neighbors to a given target vary in widely unexpected ways due to which the concept of proximity between the points in very high dimensional spaces might not be meaningful at some point. Hence the high dimensional points that failed to result in true nearest neighbors by some KNNS techniques might have to be analysed deeper. In this paper we explore some of the novel KNNS approaches that attempts to strike a synergy between their accuracy and the computation times. We then follow this with a study of a bench-marking framework that has performed an extensive evaluation of approximated nearest neighbor approaches. We also introduce a Reinforcement Learning based technique to evaluate the robustness of a KNNS algorithm against adversarial datapoints.

\section{Approaches}

Here, we describe some of the state-of-the-art nearest neighbor approaches that has proven to be effective in very high dimensional spaces. We briefly categorise the algorithms into 2 types - Graph based and non-Graph based approaches. Graph based algorithms are the ones where a graph is constructed from the data space that might be based on some approximated neighbors for each datapoint. This graph is then traversed from a suitable starting node to get the nearest neighbors. The Non-Graph based approaches cover Space Partitioning techniques, Genetic Programming etc.

\subsection{Non Graph based approaches:}

Hamid Saadatfar et al.\cite{ClusterDensity} presents a nearest neighbor technique that leverages data pruning. It builds on-top of a less improved version of data pruning based K-Nearest Neighbor (KNN) algorithm called LC-KNN \cite{LCKNN}. LC-KNN first clusters the data into smaller partitions using K-Means clustering technique, which is then followed by searching the cluster whose centroid is nearest to the given query point, there by pruning the rest of the clusters. However, the shape and the density of each of the clusters isnt't considered. This limitation is handled by  \cite{ClusterDensity} where they compute the variant spread over axes and density of the clusters by defining metrics for the same. While performing the nearest neighbor search, the algorithm picks the cluster whose max spread along an axes is below a threshold and the one that has the maximum density.

\begin{figure}[!ht]
    \centering
    \includegraphics[height=6cm]{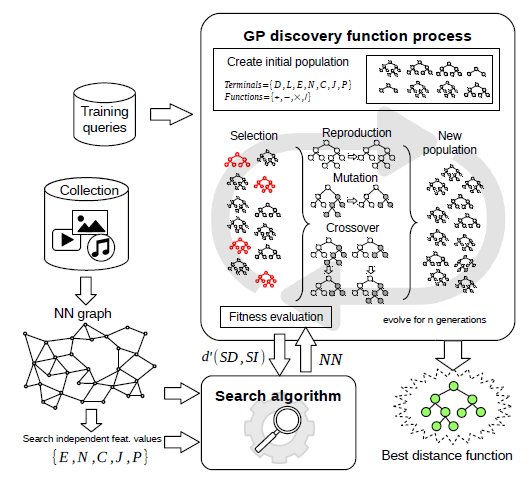}
    \caption{Architecture of GP based distance function computation for ANNS}
    \label{fig:GP}
\vspace{-3mm}
\end{figure}

\vspace{3mm}
\noindent
Javier A. Vargas Muñoz  et al. \cite{GP_2019} describes the idea of leveraging Genetic Search algorithm for the nearest neighbor search. The Genetic search algorithms eliminate set of bad solutions in every generation and allow good solutions to replicate and be modified. Each solution is assigned a fitness score based on which, it is categorised as good or bad. \cite{GP_2019} proposes an approach that constructs a nearest neighbor graph (similar to \cite{NSG}) that enables approximate nearest neighbor search. Then they use Genetic Programming (GP) to find a suitable mathematical function that combines search dependent features (distance, path length of the node to our query point) with the topological properties of the graph (edge weight, degree of vertex, etc). Each individual (discovered mathematical function) from the population is evaluated and given a fitness score (uses the recall measure as a fitness function). In the end, after repeated selection, reproduction, mutation and crossover steps, the algorithm  obtains a strong mathematical function that can in-turn be used to obtain nearest neighbors from the graph with increased speedup in comparison to other approximated approaches. The pipeline is depicted in Fig \ref{fig:GP} cited from \cite{GP_2019}- Fig.   

 \vspace{3mm}
 \noindent
 
 Michael A. Schuh et al. \cite{HybridIndexing} describes a data-space segmentation technique that reduces the candidates that need to searched to get the nearest neighbors of a query point. It is based on an existing technique called iDistance \cite{iDistance} - this technique divides the data-space into a set of partitions and a maintains a reference point for each partition. Given a query point, it then filters the partitions to search based on the proximity of the query point with the partitions, calculate the search range for each partition and retrieve the candidate points that are further refined by their true distances from the query point. \cite{HybridIndexing} extends this further and separates dense areas of the data-space by splitting the partitions into disjoint sections. It describes two ways of splitting the data-space - Global which splits the entire data-space and Local which splits the data-space of each partition into regions called segments. They describe a set of global and local heuristics for splitting the data and use $B^{+}$-tree \cite{b_plus_tree} for indexing the points and their distances to the partitions and segments. This style of data partitioning technique ensures retrieval of very few candidates that needs to be searched while determining the nearest neighbors.
 
  \vspace{3mm}
 \noindent
 Marius Muja et al. \cite{FLANN} proposes a set of approximate nearest neighbor approaches and a describe an automated configuration procedure for finding the best algorithm to search a particular data-set. They suggest the Randomized KD-tree algorithm which is similar to KD-tree in that both the approaches involve splitting the data across dimensions. However, in Randomized KD-tree, the split dimension is chosen randomly from top $N_{D}$ dimension while in KD-tree the data is split on the dimension with highest variance. The other approach is the priority search k-means tree algorithm that partitions the data points into K regions using k-means clustering. The partitioning is done recursively   until each region has smaller than K points. It then uses a priority queue to explore the branches of the constructed k-means tree while searching for neighbors of a query point. They propose an automatic selection procedure of the optimal algorithm which is highly dependent on the data dimensionality, size and structure of the data-set, desired precision. This is a cost optimization problem in which the algorithm whose parameters lead to a reduced cost is chosen to be the optimal one. Here, the cost is a combination of the search time, tree build time and tree memory overhead. Hence, depending on the application, we can get the desired optimal algorithm. They have released an open source library called fast library for approximate nearest neighbors (FLANN) that has the implementation of the techniques discussed. 
 
 \vspace{3mm}
 \noindent

  \vspace{3mm}
 \noindent
Jeff Johnson et al. \cite{FAISS} released the Facebook AI Similarity Search (FAISS) library that enables quick searching capabilities for billion-scale datasets. This in turn enabled construction of k-nearest neighbor graph on a billion high-dimensional vectors in single or multi-GPU configurations. They focus on improving the algorithm based on the metrics - speed, memory usage (RAM) and accuracy. The search is performed in parallel on multiple CPU threads or GPU. Their technique is based on quantization that maps input values (often from a continuous space) to output from a smaller finite set. Hence, they focus on approximate nearest neighbor search by compressing the data space with two levels of quantization, in particular they follow the IVFADC  indexing structure \cite{IVFADC}. They also introduced the k-selection algorithm implementation on the GPU by leveraging on the GPU's registers for performing in-register sorting and storing intermediate state. This way, the k-selection algorithm is used by the KNN to collect points with lowest distances from the query point. In the end, they have performed experimentation against variety of datasets, and also variety of algorithms and have shown satisfactory results.

\vspace{3mm}
\subsection{Graph based approaches:}
\noindent
Cong Fu et al. \cite{NSG} presents an approximated approach for the nearest neighbor search. It is based on building a Graph structure (graph index) for the given set of points where pairs of the points/nodes are connected by an edge. Given a query point, with a graph-index, one just needs to navigate this graph to determine the nodes that are closest to the query point. In particular, the technique builds a graph that belongs to a class of networks called Monotonic Search Network that consists of Monotonic Paths \cite{M_Paths}. The strategy is based on suitable selection of edges that minimises the out degree of each node. This enables faster discovery of neighbors. In order to further reduce the indexing time, while graph construction, it leverages on the idea of starting from a fixed node called the Navigating Node.
 
\begin{figure}[!ht]
    \centering
    \includegraphics[height=6cm]{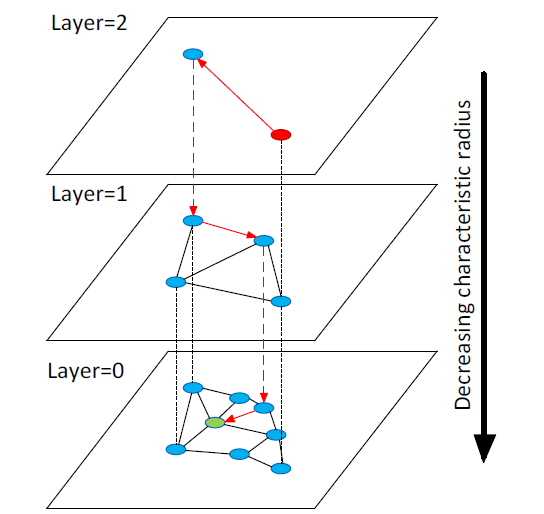}
    \caption{Idea of HNSW approach - to start the search greedily from the top-layer}
    \label{fig:HNSW}
\vspace{-3mm}
\end{figure}

  \vspace{3mm}
 \noindent
 Yu. A. Malkov et. al \cite{HNSW} proposes an approximation nearest neighbor search using Hierarchical Navigable Small World (HNSW) graphs. A small-world network is a type of mathematical graph in which most nodes are not neighbors of one another, but the neighbors of any given node are likely to be neighbors of each other and most nodes can be reached from every other node by a small number of hops or steps. The main idea of the proposed approach is to construct an approximated graph by separating the links of the nodes in the graph, according to their length scale into different layers and then performing searching in this multi-layer graph. During graph construction, for every new element, a maximum layer number $l$ is chosen according to a probability distribution and starting at the top most layer, the algorithm finds the closest neighbors to the new element $q$ and introduces bidirectional connections between them. The algorithm explicitly selects the entry point candidate in each layer from where the closest neighbors to the new element $q$ is determined. It follows certain heuristics to chose the closest neighbors of these candidates in each layer. While performing the KNN search, the algorithm starts at the top-layer, as depicted in figure \ref{fig:HNSW} and returns the candidate points determined at bottom-most layer as a part of the top-k nearest neighbors. They have compared the HNSW algorithm against the algorithms from the Euclid spaces, product quantization based algorithms and other general spaces and have shown that their performance is superior interms of accuracy.
 
\section{Evaluation}
\subsection{Existing ANN Benchmarking Framework}
\par
Here, we briefly touch upon an existing framework that does an extensive evaluation of a suite of nearest neighbor algorithms ranging from traditional ones to the state-of-the-art approximate nearest neighbor approaches. 
\par
\noindent
Martin Aumüller et al. \cite{ann_benchmark} describes an ANN-Benchmark tool for evaluating the performance of in-memory approximate nearest neighbor algorithms. They have a different notion for measuring the performance of the approximate approaches. Typically recall is measured in terms of the number of true nearest neighbors for a query point, but in this case they relax the recall measure by a threshold.
\par
\noindent
They have provided the flexibility of evaluating the algorithms against a variety of standard multi-dimensional datasets in HDF5 format. For maintaining uniformity in exploring different approaches, we have considered Twitter's vocabulary of 1.2M 100 dimensional vectors. The data-set represents the pre-trained vectors obtained from the GloVe \cite{GloVe} algorithm. They have split the data-set into train/test and provided the ground truth in the form of top 100 neighbors for each data point. We have used this dataset to study the performance of various approximate nearest neighbor approaches and also to conduct further analysis on the false neighbors of the points. 
\par
\begin{figure}[!ht]
    \centering
    \includegraphics[height=6cm]{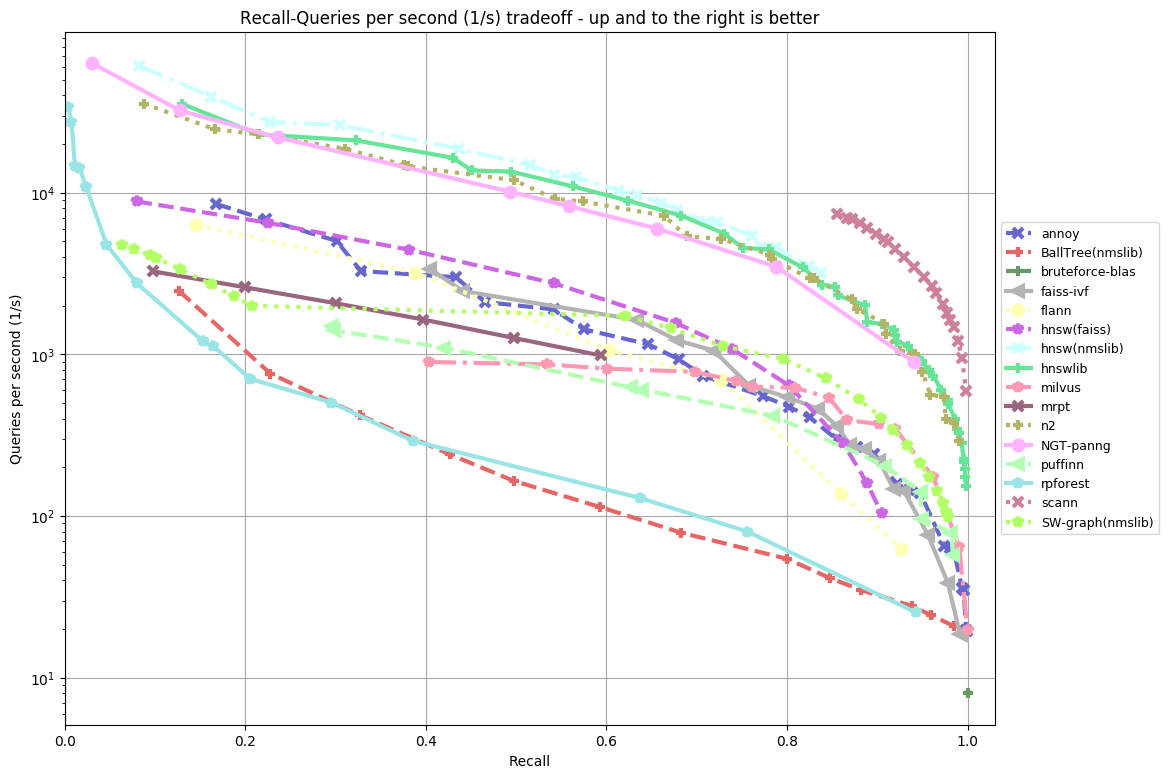}
    \caption{Performance of approximate nearest neighbor approaches}
    \label{fig:Glove}
\vspace{-3mm}
\end{figure}

\noindent
Each implementation is installed via a Docker build file that specifies the dependency installations, thus providing the flexibility to add new algorithms in the bench-marking framework. One just needs to provide a python wrapper to the algorithm, add a Docker file that specifies the installations required, and add the algorithm to the configuration file. The final execution of algorithms  and selection the datasets can be done as command line arguments of the script file. The framework picks up the specified datasets accordingly, builds an index structure for the same. This is then followed the querying the top-k neighbors for each point, where the tool also reports additional information as well. The result of the each run consists of - the nearest neighbours returned by the algorithm, time it took to find these, and their distances from the query point. The results of each run are stored in a separate HDF5 file in a directory hierarchy that encodes part of the framework’s configuration. These metrics that are a part of the resultant file can be used to visualise and compare the performances of various algorithms. The Figure \ref{fig:Glove} cited from results obtained from \cite{ann_benchmark} depicts the results obtained for various nearest neighbor algorithms - some of which were briefly touched upon in the section 2.

From the graph, we conclude that graph-based approaches shows superior performance in comparison with tree based approximate approaches. In particular the HNSW graph-based technique is relatively at its best. 
\begin{figure}[!ht]
    \centering
    \includegraphics[height=6cm]{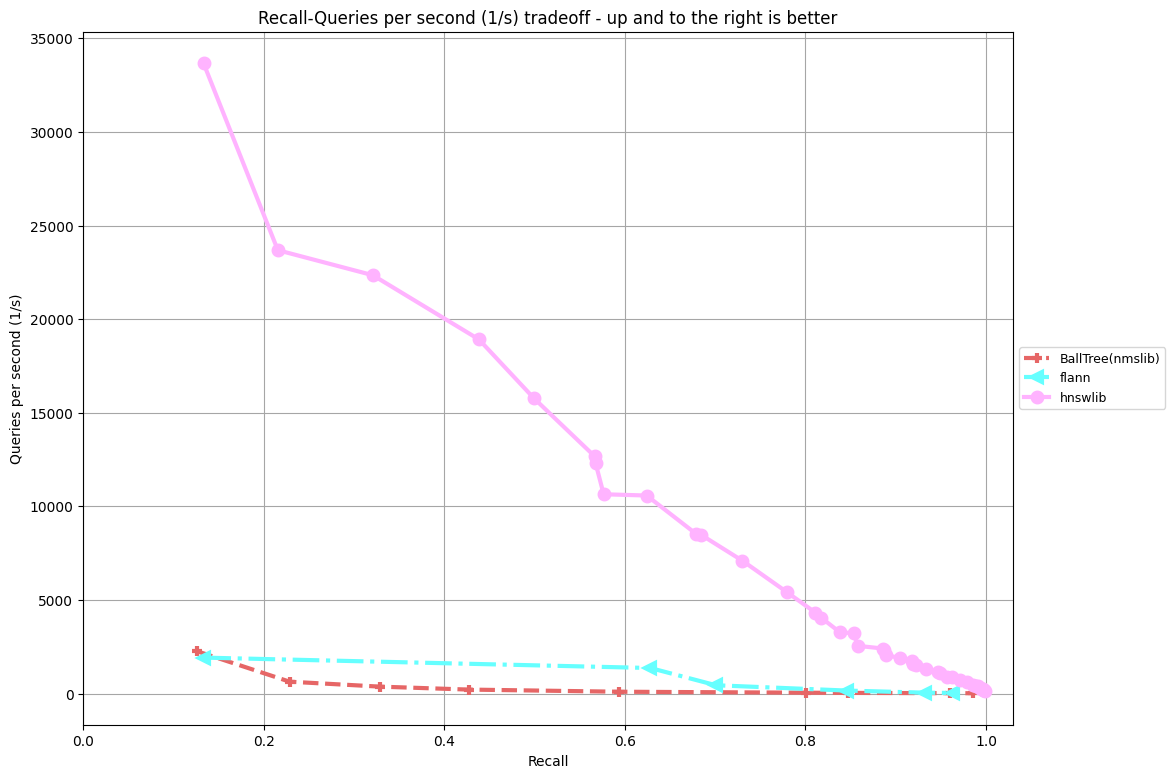}
    \caption{Performance of approximate nearest neighbor approaches computed in our machine. Although the graph looks different from the previous graph (due to different system configurations), the relationship between HNSW, FLANN and Balltree is captured. }
    \label{fig:Glove_InMachine}
\vspace{-3mm}
\end{figure}
The figure \ref{fig:Glove_InMachine} depicts the result obtained from our machine. It portrays the performance of the algorithms - graphs based HNSW, tree-based Ball Tree \cite{BallTree}.

\par
\begin{figure*}[!ht]
    \centering
    \includegraphics[height=6cm]{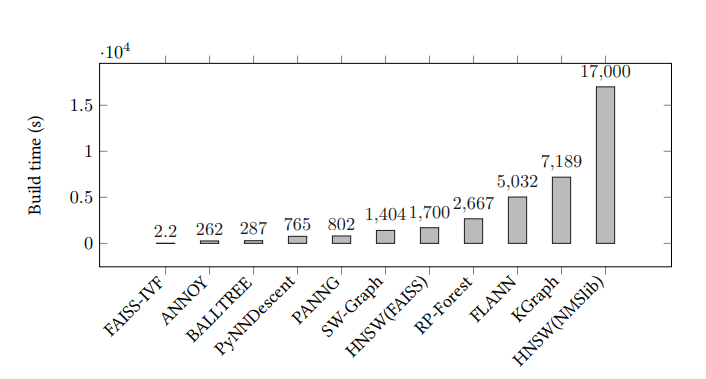}
    \caption{Index build times for different algorithms}
    \label{fig:IndexBuildTime}
\vspace{-3mm}
\end{figure*}

\noindent
The Figure \ref{fig:IndexBuildTime} shows the index build times for the GloVe dataset for different implementations. Here, we see that GPU based FAISS has the shortest index build time in comparison with building a nearest neighbor graph or a tree data structure for the dataset.
\\
We can now conclude from the above graphs that, if we are looking for highly performant algorithms in terms of accuracy of approximate neighbors, it might be better the rely on the graph-based algorithms (in particular the implementations provided by the HNSW). On the other hand if we prefer small and faster index building data structures, it might be suitable to rely on the FAISS' inverted index algorithm.
\subsection{Analysis of data points with false nearest neighbors}
In this section, we attempt to perform a deeper analysis on points which when queried using any approximate approach algorithms results in k-neighbors that do not match with the true nearest neighbors. We term such query points as False Positives (FPs). On the contrary, the ones whose approximate neighbors match with the true nearest neighbors are True Positives (TPs). We try to find if there are any evident relationships or patterns that differentiates the TPs from the FPs. We believe this kind of analysis enables one to exploit the robustness factor, explainability aspects of the approximate nearest neighbor approaches, which are difficult to interpret when dealing in very high dimensional spaces. This can in turn be useful to determine how sensitive a query point would be for a given approximate nearest neighbor algorithm.
\par
\noindent
For our exploration of different approaches, we considered test query points of size 10000 from GloVe'S 100 dimensional dataset and trained it on the FLANN approximate nearest neighbor algorithm. We then extracted the TPs and FPs by comparing with the ground truth. We have used this set of TPs and FPs for all the analysis in the future sections.\\

\par
\noindent
We initially assumed the training points to have an implicit relationship between the points that result in true neighbors and the ones that does not. Also, as we dealt with the GloVe embeddings, we suspected there could be a scenario where the model could be struggling to return true neighbors for a particular cluster of words. Hence leveraging on the dimensionality reduction techniques, we attempted to find this inherent relationship between the training points.
\subsubsection{Principal Component Analysis}\hfill\\
Principal Component Analysis (PCA)\cite{PCA} is an orthogonal linear transformation technique that transforms high dimensional datapoints to a new coordinate system such that the greatest variance component of the data lies in the first coordinate of the transformed data, second greatest variance on the second coordinate and so on.
For our use-case, we transform the test query points to a lower space and considered the first two principal components. We then visualised the classification of the TPs and FPs of the query points to determine if there existed any relationship between TPs and FPs.
\begin{figure}[!ht]
    \centering
    \includegraphics[height=6cm]{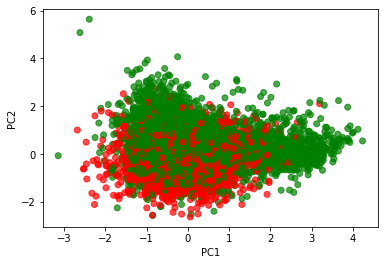}
    \caption{Dimensionality reduction via PCA}
    \label{fig:PCA}
\vspace{-3mm}
\end{figure}\\

As seen in the Figure \ref{fig:PCA} - the green dots correspond to the TPs and red dots to the FPs. From the visualisation, we can conclude that PCA might potentially shed some light on explaining the dimension that to some extent separates the TPs from the FPs. This insight may be further used in the future to model a classifier that can potentially predict if a given model will fail to compute the true neighbors for a  new incoming point.

\subsubsection{t-Distributed Stochastic Neighbor Embedding (t-SNE)}\hfill\\
t-SNE\cite{tSNE} is yet another unsupervised non-linear technique for data exploration and visualizing high-dimensional data space. While PCA tries to preserve large pairwise distances and maximize variance in the data, it can lead to poor results if there exists complex non-linear manifold structures in the data. 
\par
\noindent
t-SNE minimizes the divergence between two distributions - one that measures the similarity of points in higher dimensional space and the second one being the distribution that measures similarity of points in the lower dimensional space. In this way the t-SNE tries to find patterns in the data in the higher dimensional space and maps the same to the lower dimensional space.
\par
For our use case, we attempt to visualize the set of TPs and FPs in the query points to determine if there exists any pattern in the same. 

\begin{figure}[!ht]
    \centering
    \includegraphics[height=6cm]{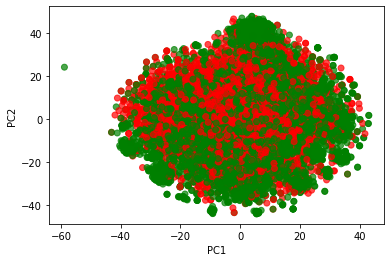}
    \caption{Dimensionality reduction via t-SNE}
    \label{fig:t_SNE}
\vspace{-3mm}
\end{figure}

As seen in the visualisation \ref{fig:t_SNE}, we cannot determine any explicit separation between the TPs and FPs points from the query points.

\subsubsection{Autoencoders}\hfill\\
Autoencoder \cite{Autoencoders} is a type of neural network that reconstructs the input as its output. The simple form of an autoencoder consists of a feed-forward neural network with input, output and one or more hidden layers. The input and the output layers has the same number of neurons/nodes. The objective of the Autoencoder is to minimize the reconstruction error of the inputs. 
\par
\noindent
The high dimensional data that goes through the input layer of the Autoencoder, gets compressed to a lower dimensional representation before getting reconstructed at the output layer. This compressed state is the latent representation of our data. The Autoencoder learns to retain all relevant information and disregard the noise from the input. This technique has been widely used to study complex relationships of very high dimensional data points by removing extraneous information of the input like processing communication signals \cite{app_latent_space_vis}, to conveniently process images, videos, manifold learning and also generate new data samples by interpolating points in the latent space. Hence, we attempted to use this approach to compress the high dimensional input points to a 2 dimensional space and visualise the latent space to hopefully gain some insights. In particular we were interested to visualise the latent representations of the TPs and FPs of the 100 dimensional Glove dataset.
\begin{figure}[!ht]
    \centering
    \includegraphics[height=6cm]{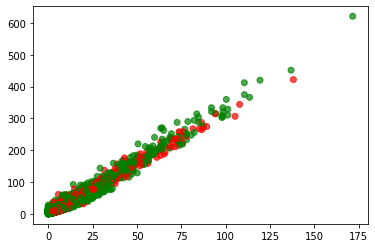}
    \caption{Dimensionality reduction via Autoencoders}
    \label{fig:AE}
\vspace{-3mm}
\end{figure}
We trained the Autoencoder model that was configured with Adam optimizer and mean-squared-error as the loss function with the query points and visualised the latent space. As seen in the Figure \ref{fig:AE} we did not get any explicit pattern between the TPs and FPs.\\

\noindent
As we chose was the GloVe dataset (which are words from Twitter), we expected to arrive at some kind of relationship between the TP and FP words. We expected the model to not be adequately trained with a semantic category of words and hence it might struggle to compute the true nearest neighbors for those words. However, although we got a fairly decent grouping between the TPs and the FPs (from PCA) we believe a deeper analysis is required to precisely differentiate the TP from the FPs.

\subsubsection{Framework for generation of adversarial examples}\hfill\\
As discussed, it is cumbersome to deal with the curse of dimensionality in high-dimensional spaces when performing the approximate nearest neighbor search. The algorithm might fail to recognise intricate patterns in the data and thus might result in neighbors that does not match with the true neighbors. We propose a framework that generates such adversarial points around a given query point. This might aid us to evaluate the robustness of KNN algorithms and conduct further analysis on the necessary quality of the dataset for a KNN algorithm.
\par
\noindent
We assume the adversarial points to have been sampled from a multivariate normal distribution that is defined by its parameters (mean vector \bm{$\mu$} and the covariance matrix \bm{$\Sigma$}). There are various ways to implement a system for adversarial learning. General Adversarial Networks\cite{GAN}, Variational Autoencoders \cite{vae}, Reinforcement Learning \cite{rl}. Considering the flexibility offered by Reinforcement Learning (RL), we chose to opt the same for our use case.
\par
\begin{figure}[!ht]
    \centering
    \includegraphics[height=6cm]{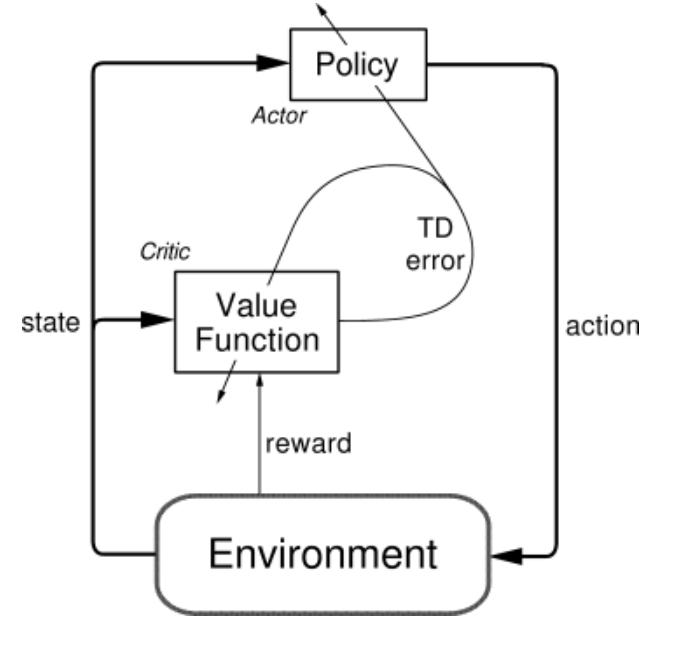}
    \caption{Actor-Critic architecture}
    \label{fig:A_C}
\vspace{-3mm}
\end{figure}

\noindent
The two main types of RL - Value Based and Policy Based methods have their own set of pros and cons. The Actor-Critic framework \cite{ActorCritic} aims to take advantage of both methods while eliminating their drawbacks. The idea is to have two components - Actor module that decides the action based on a state and the Critic module that produces the values that correspond to the maximum future rewards for the action chosen by the actor. The critic criticises the actions of the Actor network. Both these networks get better in their roles as time passes, where the Actor (Policy based)  learns the optimal policy and the Critic (Value based) learns to improve its task of evaluating the Actor. The overall architecture of the Actor-Critic is shown in fig \ref{fig:A_C} cited from \cite{A_C} Fig 1.2. For our use-case, we are using the Advantage Actor-Critic Framework where the advantage function decides how good a particular action is relative to other actions at a given state. We consider function $J(s,a)$ at the critic  to be composed of:
\begin{equation}
    J(s,a) = V(s) + A(s,a)
\end{equation}
where $J(s,a)$ is the utility function at the Critic module, $V(s)$ is the value function that indicates how good it is to be in the state $s$ and $A(s,a)$ is the advantage function that captures how good an action is at a given state while we know the value function corresponding to that state. Hence we try to make the Critic learn the advantage function as it reduces the high variance of policy networks and stabilize the model. The advantage function is given by,
\begin{equation}
    A(s,a) = r + \gamma * V(\hat{s}) - V(s)
\end{equation}
 where $r$ is the reward for that action, $\gamma$ is the discount factor and $V(\hat{s})$ is the value corresponding to new state $\hat{s}$ and $V(s)$ is the value corresponding to the current state $s$. \\

\par
\begin{figure}[!ht]
    \includegraphics[height=4cm]{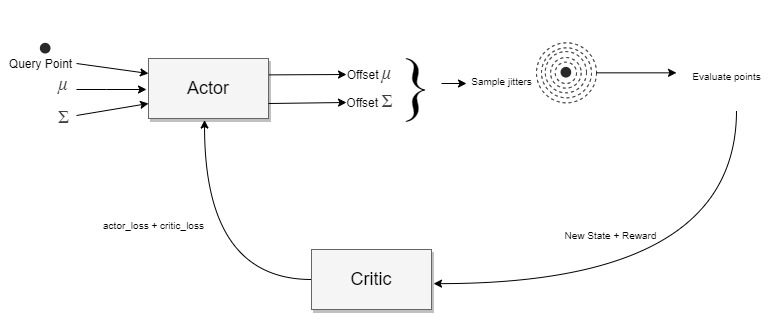}
    \caption{Design}
    \label{fig:Design}
\end{figure}

\noindent
The Figure \ref{fig:Design} depicts the design of our framework. As discussed above, we use Advantage Actor-Critic framework as the agent where - the Actor which is a neural network accepts the query point, initial parameters of the distribution. It then predicts the offset parameters (offset \bm{$\mu$} and offset \bm{$\Sigma$}) that is to be added to the initial parameters. We then sample points from this new space, and evaluate these new points/jitters. During this phase, the new points are passed onto both the true and approximate nearest neighbor approaches and evaluate the accuracy of the approximate neighbors against the true neighbors. As the magnitude of distance computations is very high, especially in case of true nearest neighbor approach, we leveraged on cuML \cite{cuML} which is a GPU based Machine Learning library. We then use the reward function:\\
\begin{equation}
    r = 100 * \log\left( \frac{\#FPs}{Sample\_Size}\right) +  CONSTANT
\end{equation}
\\
We increase the reward when the proportion of FPs in the sampled jitters is greater, which is our intent. The loss function that is back-propagated through the network is computed as:

\begin{align}
    &\delta = r + \gamma * V(\hat{s}) - V(s)\\
    &critic\_loss = \delta^2\\
    &actor\_loss = 1 - \log\left( \frac{\#FPs}{Sample\_Size}\right) * \delta\\
    &loss = actor\_loss + critic\_loss
\end{align}

where $\delta$ is the advantage function, $\gamma$ is the discount factor applied for the next state's value; We use the Critic which is an additional neural network to estimate the value for the next state $V(\hat{s})$;  $V(s)$ is the Critic network's value corresponding to the current state.

\par
\noindent

\section{Experiment and Results}
As discussed in the earlier sections, we have used 100 dimensional vectors GloVe dataset for our evaluations. We tested our RL framework for evaluation of two KNNS algorithms - FLANN and BallTree. Both the models were fitted with 10000 points out of the total training points. The fitted models were queried for top K neighbors for 10000 points (from the testing set) and the FPs and TPs were extracted by comparing the resulting neighbors with the true neighbors of the 10000 points. We use the neighbors obtained from the brute-force method using the euclidean distance as the true neighbors. 
\par
\noindent
We then used a portion of the FPs as our training set for the RL agent in order to navigate the multidimensional space. The RL agent learns to play the game by navigating the search space to find the optimal offset $\bm{\mu}$ and offset $\bm{\Sigma}$ for each of the training points. For testing purposes we configured the number of episodes for the agent to be 1 and the end goal is when all the sampled jitter points from the predicted space turns out adversarial. We chose the jitter size as 1000.
\par
\noindent
We then evaluated the robustness of the FLANN and Balltree algorithms against the adversarial points whose space is computed by our RL agent. The models computed the nearest neighbors for the jitters whose space was computed by the RL agent for each unseen point. The following metrics were computed for different values of the top-K neighbors that were to be queried :
\newpage
\begin{strip}
\centering
\includegraphics[width=\textwidth]{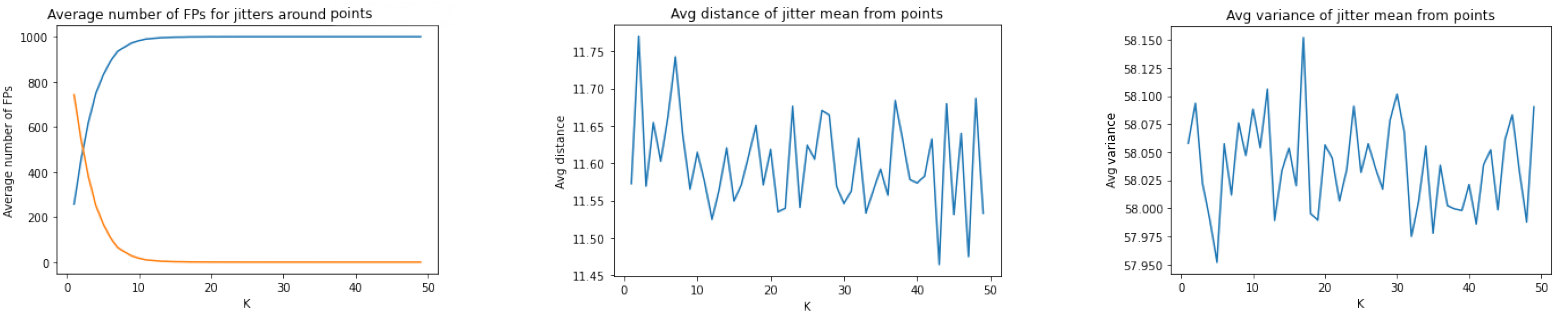}
Robustness analysis of the FLANN algorithm
 \label{fig:FLANN}
\end{strip}

\begin{strip}
  \centering
  \includegraphics[width=\textwidth]{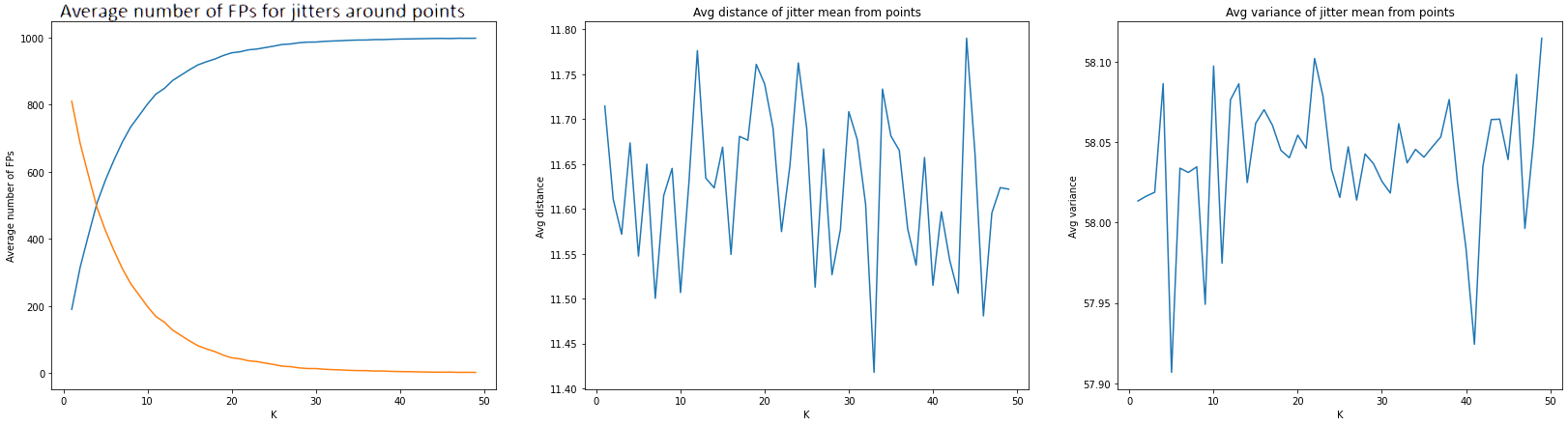}
Robustness analysis of the Balltree algorithm
 \label{fig:BALLTREE}
\end{strip}

\begin{itemize}
    \item The average number of FPs in the jitters generated around each unseen point.
    \item The average distance of jitter means from the point.
    \item The average variance of the jitters from the point.
\end{itemize}

\par
\indent

We obtained the results as shown in Figures \ref{fig:FLANN} and \ref{fig:BALLTREE}.

As seen in the figures, we realised the BallTree algorithm was relatively more robust to attacks when compared to the FLANN algorithm. This was due to the fact that, it took relatively higher values of K (when querying for top 30 neighbors) for the BallTree algorithm to turn all the jitter points to be adversarial. It can also be seen that the average distance of the jitters generated by the agent for both the algorithms was 11.6 and the average variance is 58.02. However, we believe with additional training we can get lower jitter mean and variances. The reward function can suitably be constrained to ensure the jitter means and variances stay close to their previous iterations.

\section{Conclusions and Future work}
In this paper, we discussed the complexities involved in performing K-Nearest Neighbor Search in very high-dimensional data points. We also explored some of the novel techniques that mitigated the complications caused by the curse of dimensionality. We then saw the ANN-benchmarking framework and concluded that the graph-based approaches are superior in their performances in terms of accuracy of fetching approximate neighbors. However, their index build times in comparison with the GPU based FAISS or similar non-graph approaches is longer. But this issue is a one time thing if the data space does not vary quite often. We also discussed strategies to perform deeper analysis on the points that failed in the approximate approaches and proposed a RL based framework to evaluate the robustness of the KNNS algorithms.  As part of the future work, we intend to regularise the objective function of the RL agent further to ensure the adversarial points generated stays as close to the point under consideration as possible. This can be incorporated by penalising the mean and variances of the adversarial points. As the framework is a generic one, we also intend to extend this to evaluate the robustness of other state-of-the-art KNNS approaches.




%
\clearpage
\bibliography{knns}{}
\bibliographystyle{plain}


\end{document}